\documentclass[letterpaper, 10 pt, conference]{ieeeconf}  % Comment this line out if you need a4paper

\IEEEoverridecommandlockouts                              % This command is only needed if 
\usepackage{graphics} % for pdf, bitmapped graphics files
\usepackage{graphicx}
\usepackage{verbatim}
\usepackage{color,soul}
\usepackage{multirow}
\usepackage{amsmath} % assumes amsmath package installed
\usepackage{dsfont}
\usepackage{tabu}
\usepackage{xcolor}
\usepackage{colortbl}
\usepackage[font=small]{caption}
\usepackage{subcaption}
\usepackage[%  
    colorlinks=true,
    pdfborder={0 0 0},
    linkcolor=red
]{hyperref}
\usepackage[ruled, linesnumbered, lined]{algorithm2e}
\title{\LARGE \bf
DexDLO: Learning Goal-Conditioned Dexterous Policy for Dynamic Manipulation of Deformable Linear Objects
}

\author{Sun Zhaole$^{1}$, Jihong Zhu$^{2}$ and Robert B. Fisher$^{1}$% <-this % stops a space
\thanks{$^{1}$Sun Zhaole and Robert B. Fisher are with the School of Informatics, University of Edinburgh, UK. Corresponding author:
        {\tt\small zhaole.sun@ed.ac.uk}
        }%
\thanks{$^{2}$Jihong Zhu is with School of Physics, Engineering and Technology, University of York, UK}
}

\begin{document}

\maketitle
\thispagestyle{empty}
\pagestyle{empty}

%%%%%%%%%%%%%%%%%%%%%%%%%%%%%%%%%%%%%%%%%%%%%%%%%%%%%%%%%%%%%%%%%%%%%%%%%%%%%%%%
\begin{abstract}
Deformable linear object (DLO) manipulation is needed in many fields.
Previous research on deformable linear object (DLO) manipulation has primarily involved parallel jaw gripper manipulation with fixed grasping positions. 
However, the potential for dexterous manipulation of DLOs using an anthropomorphic hand is under-explored.
We present DexDLO, a model-free framework that learns dexterous dynamic manipulation policies for deformable linear objects with a fixed-base dexterous hand in an end-to-end way.
By abstracting several common DLO manipulation tasks into goal-conditioned tasks, our DexDLO can perform these tasks, such as DLO grabbing, DLO pulling, DLO end-tip position controlling, etc. 
Using the Mujoco physics simulator, we demonstrate that our framework can efficiently and effectively learn five different DLO manipulation tasks with the same framework parameters.
We further provide a thorough analysis of learned policies, reward functions, and reduced observations for a comprehensive understanding of the framework.
%We also discuss the evidence that our framework can directly transfer into the real world.
\end{abstract}

%%%%%%%%%%%%%%%%%%%%%%%%%%%%%%%%%%%%%%%%%%%%%%%%%%%%%%%%%%%%%%%%%%%%%%%%%%%%%%%%
\section{INTRODUCATION}
\label{sec:introduction}
Deformable linear object (DLO) manipulation, e.g. ropes, cables, and rods, is widely applicable in surgical theaters, offices, textile factories, and other industries \cite{zhu2022challenges,sanchez2018robotic}.
Current research in DLO manipulation largely relies on single or dual parallel pinch grippers or end-effectors attached to a fixed end-tip of the DLO \cite{yu2022shape, chi2022iterative, lv2022dynamic, lim2022real2sim2real, yan2021learning, wang2015online}. 
However, without task-specific customization, such end-effectors cannot provide sufficient dexterity for DLO manipulation tasks like in-hand DLO sliding and DLO weight pulling (see Figure \ref{fig:approach:goal} (a), (b), and (c) respectively). Meanwhile, an anthropomorphic hand, as a versatile end-effector, has the potential to handle all the aforementioned tasks.

%such end-effectors cannot provide sufficient dexterity for manipulation tasks like living object grasping \cite{hu2023grasping}, deformable object in-hand shaping \cite{li2023dexdeform}, in-hand object re-orientation \cite{chen2022system, handa2022dextreme, andrychowicz2020learning}, re-positioning objects in space \cite{petrenko2023dexpbt,nagabandi2020deep} and rotating articulated components \cite{akkaya2019solving}. 
%Meanwhile, an anthropomorphic hand can perform the mentioned versatile tasks with dexterous manipulation.

%holds the promise of providing additional flexibility for handling all the aforementioned tasks. 

%Dexterous hands are rarely used in DLO manipulation, and meanwhile, there has been a breakthrough in the dexterous manipulation of rigid objects.

%However, dexterous manipulation of DLOs is missing in the literature.
There are three common practices in the above-mentioned \textit{traditional} DLO manipulation methods: 1) quasi-static DLO manipulation with a near zero velocity, 2) fixed grasp on the DLO, and 3) customizing end-effectors to execute a specific DLO task.
Compared to a two-finger gripper or DLO end-tip fixed end-effector, an anthropomorphic hand can avoid relying on these requirements during DLO manipulation, as shown in Figure \ref{fig:introduction}, where the anthropomorphic hand grabs a DLO to fetch its end-tip.
The major technical challenges of dexterous DLO manipulation are:

\begin{figure}[tbp]
  \centering
  \includegraphics[width=0.48\textwidth]{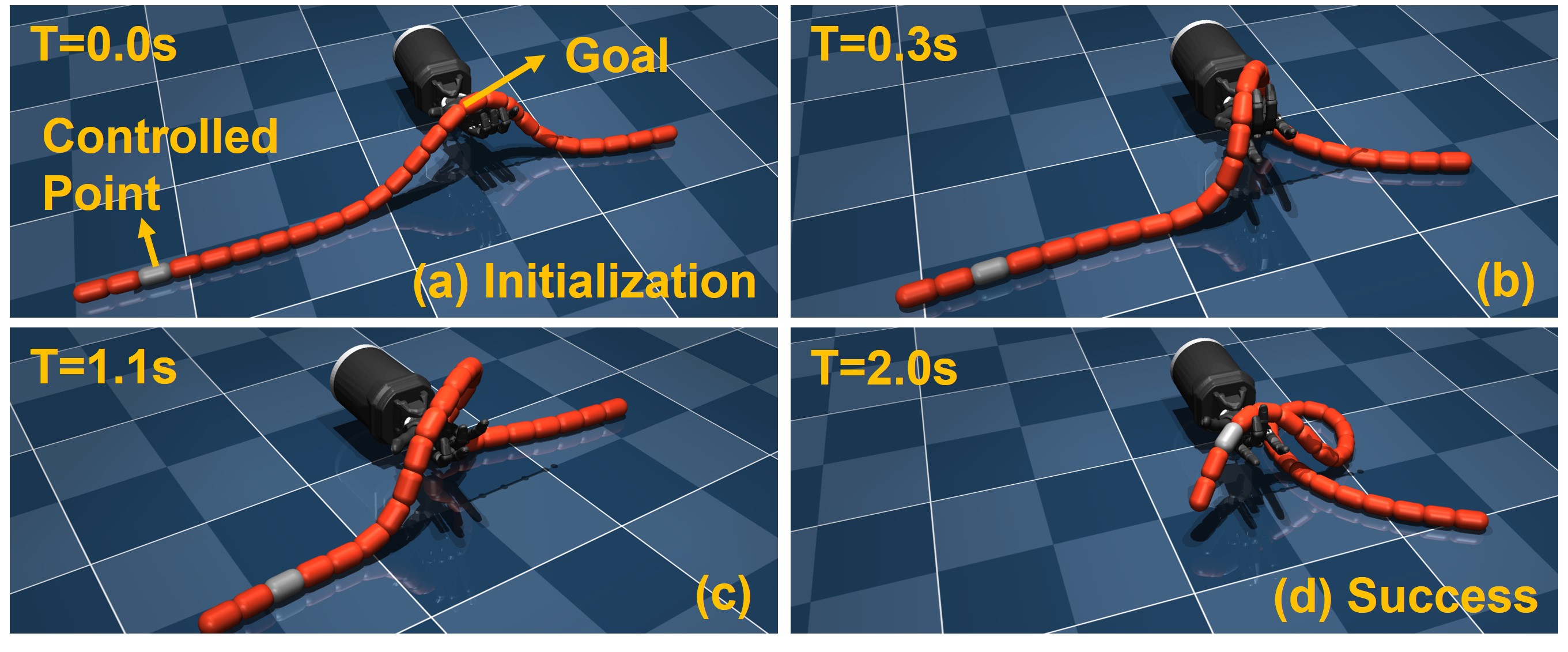}
  \caption{An example of goal-conditioned dexterous manipulation of a deformable linear object. The \textit{controlled point} (the grey segment) on the DLO is manipulated to minimize the distance to its \textit{goal position} (the hand palm) and finally reach the \textit{goal position} using a base-fixed dexterous hand.}
  \label{fig:introduction}
  \vspace{-7mm}
\end{figure}

\begin{itemize}

\item \textbf{Dynamic manipulation}.
Previous continuous control methods mostly manipulated a DLO in a quasi-static state, assuming the DLO's velocity is near zero \cite{yu2022shape}, and this assumption excludes some scenarios with non-zero velocities and higher manipulation speeds.
Though handling the complexity of dynamics during real-time manipulation of the DLO is very difficult, dexterous manipulation of DLOs can benefit, making it more adaptable to different tasks.

%Handling the complexity of dynamics in real-time manipulation of the DLO can be very difficult but highly beneficial.

\item \textbf{Changing grasping positions}.
Preventing the DLO from slipping or falling from the hand during changing grasping positions is challenging for parallel jaw grippers.
Dexterous hands offer a unique advantage for this task without having to place and regrasp the DLO, which is often unavailable without a supporting surface, e.g. when picking and placing a rope on a table \cite{yan2021learning}. 

\item \textbf{General end-effectors}.
Different end-effectors are often used for different DLO manipulation tasks, including specialized end-effectors, e.g., a gripper with tactile sensors \cite{she2021cable} in DLO following\footnote{Here, `following' is defined to mean sliding the DLO through the hand.} and a specially designed gripper for rolling flat cables \cite{chapman2021locally}.
Dexterous hands are general-purpose end-effectors suitable for various rigid object manipulation tasks.
Chen et al. \cite{chen2022system} have shown that reorienting long and thin rigid objects in hand is difficult.
Thus, it is even more difficult to use dexterous hands to manipulate DLOs, which are \textit{highly deformable} long and thin objects and have high DoFs. 

%It is also complicated to design a single framework to achieve many DLO manipulation tasks that can be solved using a dexterous hand.

\end{itemize}

Some works have already explored one or two aspects, e.g., DLO following to change grasping positions within the gripper with tactile sensors \cite{she2021cable} or specialized end-effectors \cite{chapman2021locally} and DLO dynamic whipping with a fixed grasping position \cite{chi2022iterative, lim2022real2sim2real, zhang2021robots}. However, none have systematically studied the dexterous manipulation of DLOs to address all three challenges, as discussed in Section \ref{sec:related}.

To address the mentioned challenges, we introduce \textbf{DexDLO}, a reinforcement learning based framework for DLO dexterous manipulation, together with a pose-regularized reward function.
DexDLO is an end-to-end framework designed to control a dexterous hand to minimize the distance between a selected point \textbf{X} on the DLO and a goal position \textbf{G} in a dynamic way without explicitly regrasping the DLO or moving the hand base (see Figure \ref{fig:introduction} for an example).
%We show that using DexDLO, the hand can grab the DLO's one end-tip with dexterous manipulation in 2 seconds .
Without changing any structure or parameters of DexDLO, it can solve a series of goal-conditioned DLO manipulation tasks by formulating these tasks into goal-conditioned forms.%, as stated in Section \ref{sec:approach:problem}.

We evaluated the framework on five different goal-conditioned tasks and summarize our contributions as follows:

%By evaluating the framework on five different goal-conditioned tasks, we show the proposed framework's capability and adaptability (see the experiment results in Section \ref{sec:experiment}, including the experiment setup, experiment evaluations, and ablation studies). 

%Although the experimental results presented in this paper are based on simulation, Section \ref{sec:discussion} discusses why DexDLO can be implemented in the real world.

%The paper's  contributions are:

\begin{itemize}
    %\item We summarize and define a series of DLO manipulation tasks into a goal-conditioned task.
    %\item \textcolor{red}{something about a general framework for solving a number of Deformable Linear Object manipulation tasks.}
    \item A general formulation of goal-conditioned deformable linear object (DLO) manipulation as applicable to several challenging DLO manipulation tasks (see Section \ref{sec:approach:problem}).
    \item The first general framework that learns manipulation policies that can perform a series of goal-conditioned dexterous DLO manipulation tasks (see Section \ref{sec:approach:framework} for the framework and Section \ref{sec:exp:results} for evaluation).
    \item An analysis of removing certain observations and reward terms on the learned policy, finding that the observation space can further be reduced and pose-regularized reward terms are essential for training (see Section \ref{sec:exp:reward}).
    
    %\item We discussed the gap between the simulation and the real world to provide evidence that the policies learned in our framework can transfer into the real world.
\end{itemize}

%The paper is structured as follows. We further discussed DLO manipulation of various tasks and goal-conditioned manipulation in DLO manipulation and dexterous manipulation, respectively, in Section \ref{sec:related}. 
%In Section \ref{sec:approach}, we introduce our definitions of several goal-conditioned DLO manipulations in Section \ref{sec:approach:problem} and our DexDLO framework in detail in Section \ref{sec:approach:framework}. 
%We show our experiments in Section \ref{sec:experiment}, including the experiment setup, experiment evaluations, and ablation studies. 
%In Section \ref{sec:discussion}, we discuss the evidence that our DexDLO can be implemented in the real world.

\section{RELATED WORK}
\label{sec:related}
%\textbf{Dexterous Manipulation.} By using dexterous end-effectors, usually multi-finger hands, the robot can perform dexterous manipulation of objects, including rigid object manipulation, like object grasping \cite{xu2023unidexgrasp}, reorientation \cite{chen2022system}, and solving a Rubis's cube \cite{akkaya2019solving}. Meanwhile, there is another work studying deformable object shaping from human demonstration, DexDeform, proposed by Li et al. \cite{li2023dexdeform}.

\textbf{Deformable linear object (DLO) manipulation} has long been studied with various end-effectors on different tasks, including: 1) shape control by using a parallel-jaw gripper for pick-and-place shape control of the DLO \cite{yan2021learning, seita2021learning}, and fixing two end-tips of the DLO to do 3D shape control \cite{yu2022shape, lv2022dynamic}, 2) tying knots with dual grippers \cite{suzuki2021air, saha2007manipulation} and a three-finger hand \cite{yamakawa2007one}, 3) untangling DLOs lying on a table with discrete pulling actions by two grippers \cite{viswanath2022autonomously}, 4) DLO insertion into a hole, by grasping the DLO with a gripper \cite{chebotar2019closing, wang2015online}, 5) whipping, by fixing one end of the DLO and fixing the other end to the robot arm \cite{chi2022iterative}, 6) following/sliding, moving along the DLO from one part to another (usually the end-tip), which requires a relative movement between the end-effector and the DLO, by either using a specially designed gripper \cite{chapman2021locally} or tactile information \cite{she2021cable}. 

In the referenced literature, frameworks are often designed for specific tasks and have limited versatility. However, by consolidating the formulations of various DLO manipulations into a goal-conditioned approach, we introduce a model-free, end-to-end framework based on reinforcement learning. The proposed single framework can handle a wide range of DLO manipulation tasks without needing customization for each specific task.

%\textcolor{red}{We need some sort of conclusion, eg: These tasks have all been previously achieved using different frameworks, whereas here we show that all can be achieved within the single DexDLO framework.}

\textbf{Goal-conditioned manipulation} achieves different goals in particular manipulation task scenarios without considering intermediate partial-solution states \cite{chi2022iterative}.
It is a popular approach to dexterous manipulation and deformable object manipulation. 
Previous works focused on dexterous manipulation.
These include in-hand object reorientation to a goal pose \cite{chen2022system, handa2022dextreme, petrenko2023dexpbt}, solving Rubik's cube \cite{akkaya2019solving}, tossing cubes to goal positions \cite{petrenko2023dexpbt}, using tools to finish goal tasks \cite{chen2022towards, rajeswaran2017learning}.
These works also include deformable object manipulation, including rope whipping to hit a goal in the air \cite{chi2022iterative, zhang2021robots, lim2022real2sim2real}, shape control \cite{yan2021learning, yu2022shape, seita2021learning}, insertion \cite{chebotar2019closing}, and DLO following \cite{she2021cable}.
A gap still exists in %researching 
the integration of goal-conditioned DLO manipulation with dexterous manipulation.
%\textcolor{red}{paragraph needs some conclusion statement about the limitations of the previous approaches}

\section{APPROACH}
\label{sec:approach}
This section formulates goal-conditioned DLO Manipulation and then briefly describes the DexDLO framework.

\subsection{Goal-conditioned DLO Manipulation}
\label{sec:approach:problem}

\begin{figure}[tbp]
  \centering
  \includegraphics[width=0.48\textwidth]{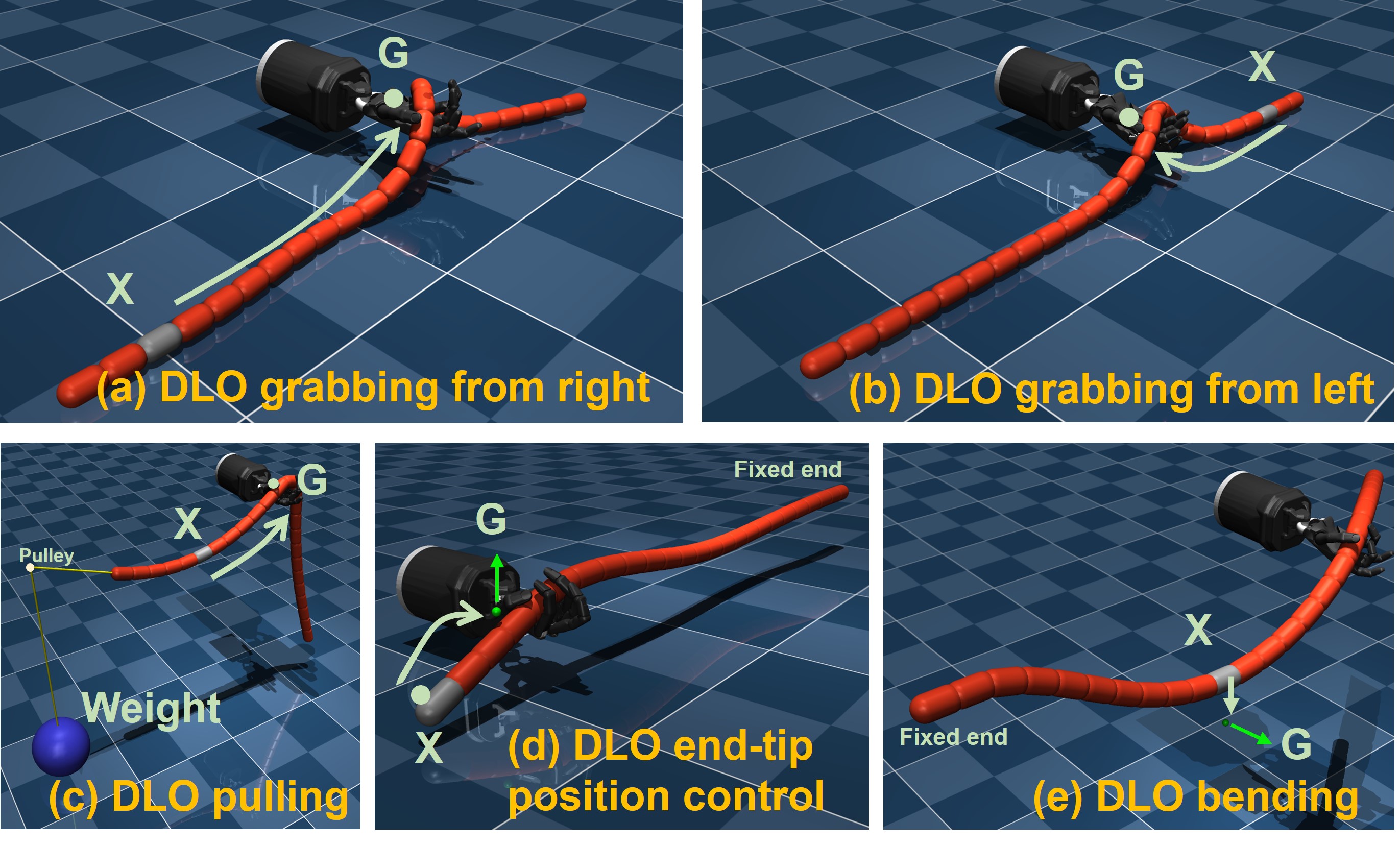}
  \caption{Five example goal-conditioned DLO manipulation tasks. The grey segment on the DLO is the controlled point $X$, and the goal $G$ is at the hand palm or highlighted as the green motion direction. (a) and (b) are two DLO grabbing tasks but from different directions. (c) is DLO pulling, where the weight is hung through a fixed pulley and connected to the end-tip of the DLO. (d) is DLO end-tip position control where the hand is fixed close to the end-tip, and another end-tip is fixed. (e) is DLO bending where the hand grasps near to one end-tip, another end-tip is fixed, and the middle point of the DLO is the controlled point.}
  \label{fig:approach:goal}
  \vspace{-3mm}
\end{figure}

Define a DLO as $n$ connected keypoints $P_0, P_1, ..., P_{n-1}$.
All $P_i\in R^3$ consist of a point position along the DLO.
A goal-conditioned policy is a point-to-point policy that moves a designated point $X = P_i \in R^3$ $(i = 0,1,2..n-1)$ to a specified goal position $G \in R^3$.
%A goal-condition policy can be described as a point-to-point policy by moving $X \in \{P_0, P_1, ..., P_{n-1}\}$ to a goal position $G \in R^3$ and minimizing the distance $||X - G||_2$.
In each task, $X$ is initialized to be a specific keypoint and remains fixed during the given task.
We set the threshold distance of  success to $d$ where $||X-G||_2 < d$ indicates a success.
%The notations are illustrated in Figure \ref{fig:approach:goal}. 

%More simulation details are given in Section \ref{sec:exp:setup}.
We introduce five different DLO tasks and set $n=25$.
%\textcolor{red}{In some places you refer to six or five tasks. check for consistency.}

\textbf{DLO Grabbing.}
The goal is to grab a specified point $X = P_i$ on
%certain place of 
the DLO by the base-fixed shadow right hand.
The DLO is initially placed on the upper area of the hand palm and falls freely, where $P_0$ is on the right-hand side and $P_{24}$ is on the left-hand side.
The hand needs first to grasp the DLO and use its motions to move the point $X$ into the goal position $G$ in the hand palm without dropping the DLO.
We have two DLO grabbing tasks by setting $X$ to be either $P_2$ or $P_{22}$, which the hand needs to then slide from the right to the left (or the opposite way), as shown in (a) and (b) of Figure \ref{fig:approach:goal}. 
%We explore if the hand can grab the DLO's end-tip successfully.
We set the threshold $d=0.08m$.
These two tasks are inspired by previous cable following work \cite{she2021cable} and the special end-effector designed to grab the DLO \cite{chapman2021locally}.

\textbf{DLO Pulling.}
The DLO is pulled over a pulley with a weight attached to the DLO at the right-hand side of the hand, as shown in (c) of Figure \ref{fig:approach:goal}.
The hand must pull the DLO to counter the external force caused by gravity on the weight.
We set $X$ to $P_4$ and $G$ to the hand palm.
The pulley is abstracted to be an anchor fixed at a certain place connecting to $P_0$. By pulling the DLO, $P_5$ moves to $G$, and the weight can be lifted.
We set the maximum weight as $0.4kg$.
We set the threshold $d=0.08m$.
%We explore if the hand can pull the DLO with the weight.

\textbf{DLO End-tip position control.}
The controlled keypoint $X=P_0$ of the DLO is manipulated into the goal position $G$ in 3D space.
We assume that the end-tip of the DLO is grabbed, and the hand should further manipulate it to a certain position, which is a simplified setting of the DLO insertion task \cite{chebotar2019closing}.
The goal position $G$ is sampled in a spherical space whose radius equals $0.1m$. The center of the spherical goal space is the middle point between the initial position of $X$ and the hand palm.
The hand is fixed $10cm$ beneath $P_{7}$. The left-hand side end-tip $P_{24}$ is fixed.
We set the threshold $d=0.05m$.
%We want to know if the hand can learn to manipulate the DLO end-tip to reach different target places. 

%\textbf{DLO End-tip position trajectory following}

\textbf{DLO Bending.}
Although DLO shape control with one robot arm or a dual-arm system using fixed end-effectors, like grippers has been well-studied previously  \cite{yu2022shape, lv2022dynamic}, the research assumed that the grasping position on the DLO is fixed and the shape is controlled by moving the gripper.
This problem is solved from a different perspective in a simplified version.
We assume the right-hand side end-tip $P_{0}$ of the DLO is fixed.
% and manipulate the DLO with non-static grasping positions and a base-fixed hand.
The controlled keypoint $X=P_{11}$ near the middle of the DLO is bent at a goal position $G$.
The hand is fixed $10cm$ beneath the initial position of $P_{20}$ near the left-hand side end-tip $P_{24}$.
Since the reachable places of the middle keypoint are unknown, we only consider if the height of $G$ can be reached by $X$, 
%\textcolor{red}{[$\leftarrow$should this be X?]}
where $G$ is $0.1m$ to $0.4m$ beneath the initial position of $X$.
The bending is successful once $X$ reaches the height of $G$. 
We set the threshold $d=0.05m$.
%We explore the conditions under which DexDLO can achieve this.

\subsection{DexDLO Framework}
\label{sec:approach:framework}

\begin{figure}[thbp]
  \centering
  \includegraphics[width=0.48\textwidth]{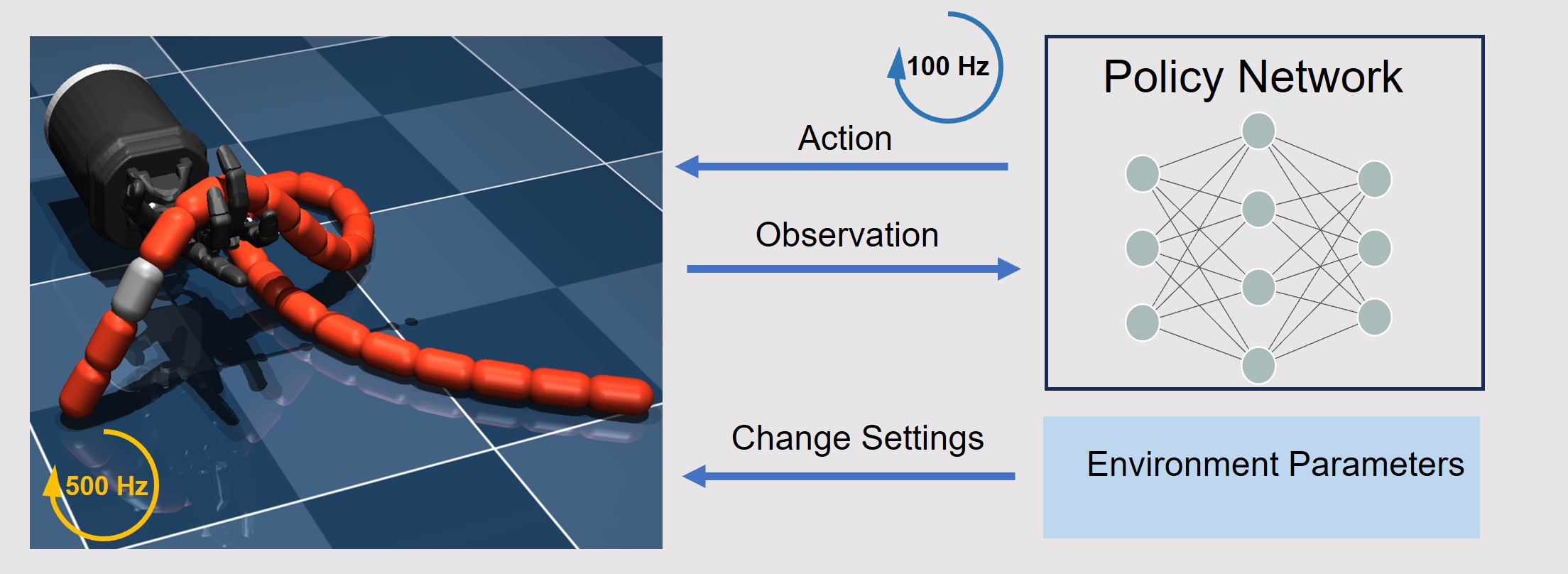}
  \caption{Policy training. The policy network predicts actions at 100Hz, while the hand controller runs at 500Hz to execute actions.}
  \label{fig:approach:framework}
  \vspace{-3mm}
\end{figure}

To solve the problem of dexterous manipulation of DLOs, we use model-free reinforcement learning (RL).% and formulate this control problem as a Markov Decision Process (MDP).
We use Proximal Policy Optimization (PPO) \cite{schulman2017proximal} to learn a policy $\pi_\theta (o)$, which takes observations $o \in O$ as input and predicts actions $a \in A$, to maximize the expected episodic discounted sum of rewards $max_{\theta}$ $E_{\pi_{\theta}}(\sum_{t=0}^{T} \gamma^t r(s_t, a_t))$.
We use the PPO implementation from Stable-Baseline3 \cite{stable-baselines3} with the \textbf{default} setting without tuning any hyper-parameters except the MLP structure, which has 1024-512-128 nodes for each layer with ReLU as the activation function.
Figure \ref{fig:approach:framework} shows the policy training model.

\textbf{Observation Space $O$}.
The observation space is given in Table \ref{table:observation}. The dimension of the observation $O_t$ is 393 at each timestep $t$.
A four-frame stack with $\{O_t, O_{t-1}, O_{t-2}, O_{t-3}\}$ is the policy network input so that the temporal dynamics can be modeled.

The observation space is redundant.
Removing some observations while maintaining similar performance is possible.
These removable observations include 
1) palm-DLO position and vector,
2) some DLO keypoint positions, and
3) hand joint velocities.
Simplification is discussed in Section \ref{sec:exp:reward}. 
%\textcolor{red}{I simplified the above paragraph and replaced Cable by DLO in the table for consistency. Please confirm/correct.}

\begin{table}[thbp]
\centering
\caption{Observation Space}
\label{table:observation}
\begin{tabular}{cc|cc}
\hline
Observation                                                              & Dim  & Observation                                                                & Dim \\ \hline
Hand joint positions                                                      & 24   & Hand joint velocities                                                        & 24  \\ \hline
DLO keypoint positions                                                   & 3$\times$25 & Hand fingertip positions                                                    & 3$\times$5 \\ \hline
Hand palm position                                                       & 3    & Goal position                                                              & 3   \\ \hline
DLO to palm vectors                                                     & 3$\times$25 & DLO to palm distances                                                     & 25  \\ \hline
\begin{tabular}[c]{@{}c@{}}DLO to fingertip \\ distances\end{tabular}   & 5$\times$25 & \begin{tabular}[c]{@{}c@{}}Target keypoint to goal\\ distance\end{tabular} & 1   \\ \hline
\begin{tabular}[c]{@{}c@{}}Target keypoint to goal\\ Vector\end{tabular} & 3    & Actions                                                                     & 20  \\ \hline
\end{tabular}
\end{table}
    
\textbf{Action Space $A$}.
The shadow hand contains 24 DoFs and 20 actuators that control the joint angles.
The actor network predicts a 20-dim vector as the target joint angles. 
The shadow hand's joints move to the target joint angles using a PD controller.
Following Chen et al. \cite{chen2022system}, we add a joint angle change limit of 0.4 rad per timestep for each joint by clipping actions outside of the limit to avoid rapid motions.

%Following OpenAI's practice \cite{rajeswaran2017learning, akkaya2019solving} of discretizing the action space into 11 bins of equal size, the agent predicts a 20-dim action vector $a \in \{\pm1, \pm0.8, ... \pm0.2, 0\}_{20}$ proportional to the ranges of joint angles.
%The actions are relative to current joint angles.

\begin{figure}[thbp]
  \centering
  \includegraphics[width=0.48\textwidth]{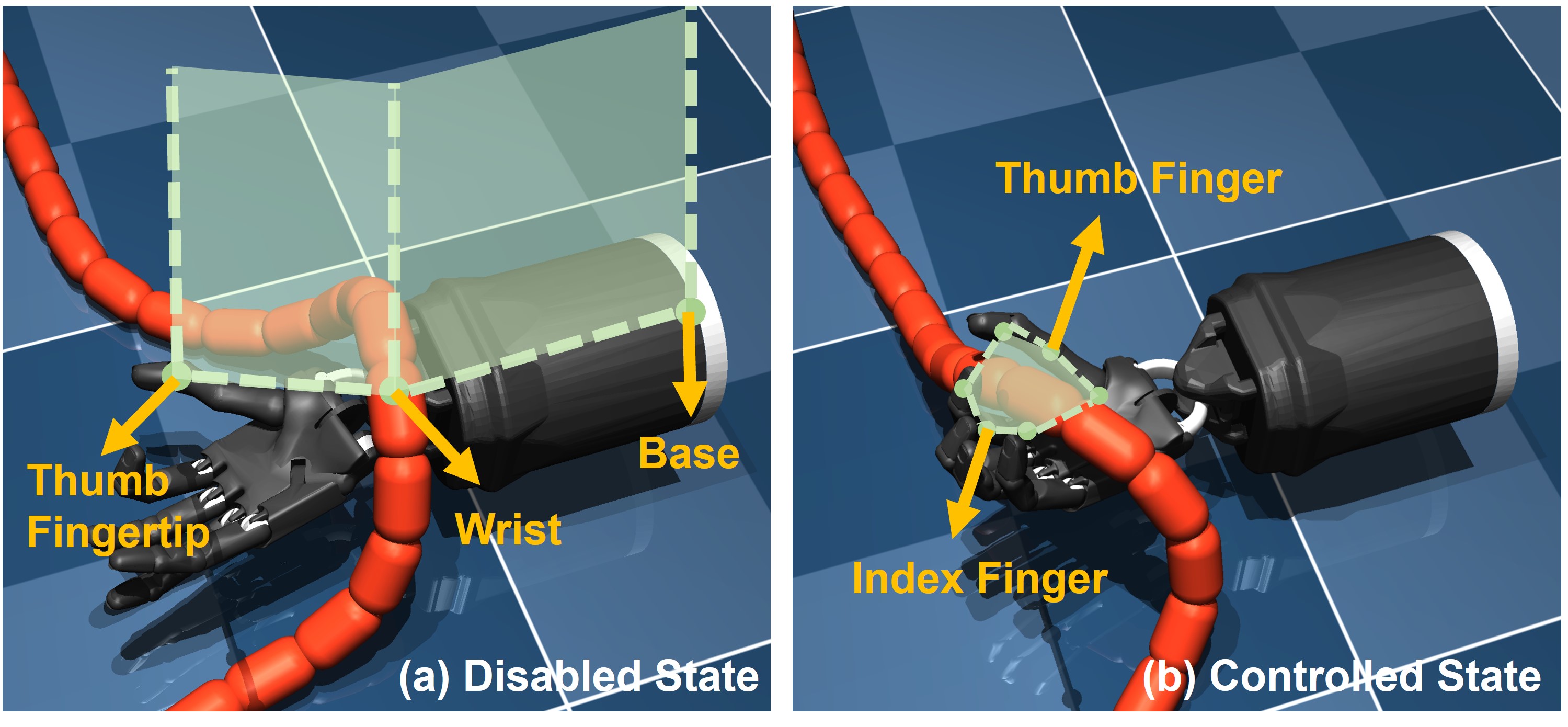}
  \caption{Disabled state and controlled state. (a) The  disabled state is set when the DLO penetrates the green area formed by the upper plane between the thumb fingertip, the wrist, and the base. (b) The controlled state is set when the DLO penetrates the green area, which consists of multiple triangles formed by the thumb and the index finger,}
  \label{fig:approach:thumb}
  \vspace{-3mm}
\end{figure}
\textbf{Reward Function $R$}. The reward function is important for guiding the agent to learn complicated manipulation skills. 
An intuitive, sparse goal-reaching reward may not provide enough guiding information when the search space is too large or complicated.
For example, DexPBT used a multiple-stage reward function to learn their policy \cite{petrenko2023dexpbt}.
For all tasks, we use a unified reward function:

\begin{align*}
    %r &= r_{reach} \times r_{controlled} + r_{disabled} + r_{success} + r_{failure},
    r &= r_{reach} + r_{disabled} + r_{success} + r_{failure},
\end{align*}

\noindent where the four reward terms are:

\textit{1. Goal-reaching-based reward $r_{reach}$}: 
\begin{align*}
    r_{reach}(\Delta  \geq 0) &= \lambda_1 \Delta  \mathds{1}_{controlled} \\
    r_{reach}(\Delta  < 0) &= \lambda_1 \Delta .
\end{align*}

The reaching reward is given based on the change of the distance $\Delta  = ||X_{t-1} - G||_2 - ||X_{t} - G||_2$ between each timestep.
The positive reaching reward is applied only if the DLO is in the controlled state, meaning $\mathds{1}_{controlled} = 1$. 
A DLO held by the hand with the thumb and index finger is in the controlled state, as shown in Figure \ref{fig:approach:thumb}b.
The controlled state encourages the DLO to be grasped in the closure between the thumb and index fingers, so that the thumb can perform its control action.
We choose $\lambda_1 = 100$, where the length of the DLO is around 1.45m.

\textit{2. Pose-regularized reward $r_{disabled} = \lambda_2 \mathds{1}_{disabled}$}.
The hand can barely control the DLO when the DLO is %manipulated to the
positioned above the wrist area.
To avoid this location, we set an uncontrollable state as $\mathds{1}_{disabled}$ and penalize the agent when the hand-DLO system reaches this state.
$\mathds{1}_{disabled} = 1$ indicates the hand-DLO system is in the disabled state. 
Like the controlled state, the DLO is in a disabled state when it penetrates the vertical surface from the thumb to the wrist and the base, shown in Figure \ref{fig:approach:thumb}a.
We use $\lambda_2 = -1$.

%\textcolor{red}{[These rewards are formulated as hard rewards. Future research could explore softer rewards that approach 1 as the target plane/position is approached.]}

\textit{3. Success reward $r_{success} = \lambda_3 \mathds{1}_{success}$}.
When the target point on the DLO reaches the goal position within the threshold distance $d$, as mentioned in Section \ref{sec:approach:problem} for each task, for more than $5$ timesteps without leaving, then this episode is a success. 
We use $\lambda_3 = 100$.

\textit{4. Failure reward(penalty) $r_{failure} = \lambda_4 \mathds{1}_{failure}$}.
There are two failure cases: 1) the DLO dropping away from the hand and 2) staying in the disabled state for over $50$ timesteps.
More specifically, a dropping failure is when the distance from the nearest DLO keypoint to the palm is further than $10cm$, or the highest DLO keypoint is less than $5cm$ above the floor.
We use $\lambda_4 = -100$.

\section{EXPERIMENTS}
\label{sec:experiment}

This section presents the experiment setup, performance evaluation, learned policy analysis, and ablation studies on observation and reward designs. 

\subsection{Experiment Setup}
\label{sec:exp:setup}

\begin{figure*}[th!]
  \centering
  \includegraphics[width=0.98\textwidth]{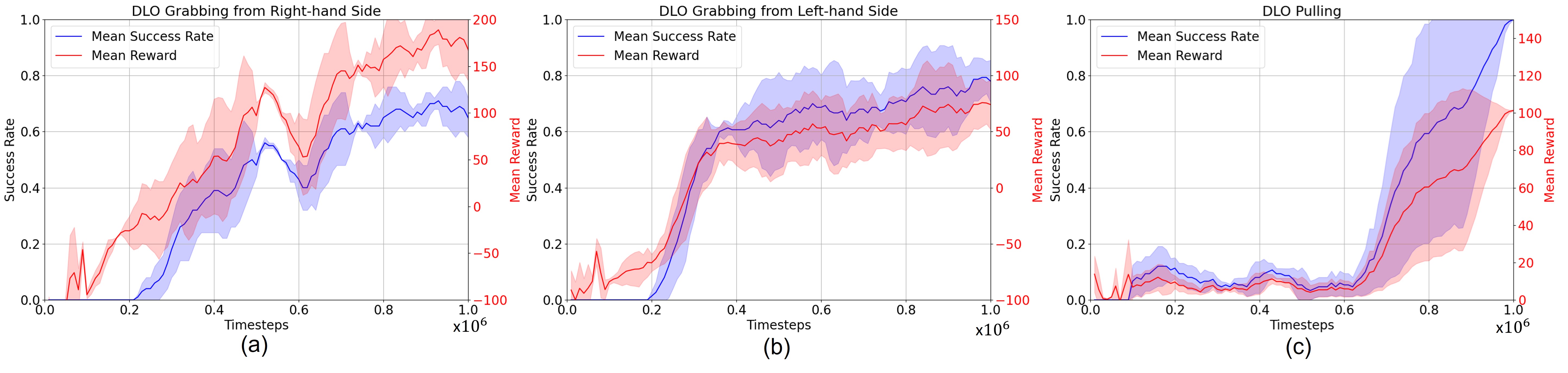}
  
  \caption{Training curves for (a) DLO grabbing right, (b) DLO grabbing left, and (c) DLO pulling. \textcolor{blue}{Blue} is the success rate, and \textcolor{red}{Red} is the mean reward per episode. The shaded area indicates the standard deviation of repeated policies.}
  \label{fig:exp:evaluation:top3}
  \vspace{-3mm}
\end{figure*}

We use MuJoCo \cite{todorov2012mujoco} as our simulation environment and a simulated Shadow Right Hand as our dexterous hand, which is one of the most commonly used dexterous hands in previous works \cite{chen2022system, rajeswaran2017learning, akkaya2019solving}. The Shadow Hand is an anthropomorphic robotic hand with 20 actuators and 24 DoFs.
The DLO is modeled by 25 capsules connected with spherical joints.
We only ran a single environment without parallelism in MuJoCo, and all experiments used a CPU having a clock speed of 3.5 GHz.
The maximum episode length is normally 1500 steps (DLO pulling uses 3000 steps for the episode length).
Each agent was trained with $1\times10^6$ steps, which takes around 5 hours, and we evaluated the agent every $1\times10^4$ steps.
Each experiment is repeated 3 times, and the training curves in Figures \ref{fig:exp:evaluation:top3} and \ref{fig:exp:evaluation:lower2} show the mean and variance.
%We focused on evaluating and analyzing DLO grabbing from two directions and DLO pulling, and we tested DexDLO's availability of DLO bending and DLO end-tip position control.
% https://github.com/deepmind/mujoco/issues/203

\begin{table}[t]
\centering
\caption{DLO parameter randomization range. The stiffness unit is Nm/rad.}
\begin{tabular}{cc}
\hline
DLO parameter & Range           \\ \hline
Radius (mm)   & $20\pm3$    \\
Length (m)    &  $1.45\pm0.10$   \\ %\hline
Stiffness of Soft DLO     & $0.020\pm0.005$   \\
Stiffness of Medium DLO   & $0.2\pm0.05$   \\ 
Stiffness of Stiff DLO    & $2\pm0.5$   \\
Coefficient of Friction & $0.5\pm0.2$  \\
Mass (g)      & $250\pm25$     \\ \hline
\end{tabular}
\label{table:approach:DLO_param}
\end{table}

\begin{table}[t]
\centering
\caption{Hand-DLO Placement.}
\begin{tabular}{cc}
\hline
Hand-DLO Placement& Range           \\ \hline
Hand Base Position (m)  & $[0\pm 0.04, 0\pm0.04, 0\pm0.04]$    \\
DLO Initial Angles (rad) & $0 \pm 0.02$\\ \hline
\end{tabular}
\label{table:approach:env_setup}
\end{table}

We adjust different environment parameters during training, which include three aspects: 1) DLO parameters, 2) environment setup, and 3) observation and action noise.

\textbf{DLO parameters}. We adjust the radius, length, three types of stiffness, friction, and mass, shown in Table \ref{table:approach:DLO_param}.
%We only use one type of stiffness in each experiment.
Adjusted DLO parameters are uniformly sampled from the range after 10 successes.
Starting from the range center, we use an adaptive scheme to update the DLO parameter randomization, depending on the number of successes. 
As successes accumulate, the sampling range extends linearly until it spans the full range at 100 successes, where the full range is in Table \ref{table:approach:DLO_param}. 
%\textcolor{red}{[$\leftarrow$ I don't understand this. Zhaole: I have written the in-detailed version.]}
For the two DLO grabbing tasks and DLO bending, we use DLOs with medium stiffness. For DLO pulling, we use a soft DLO.
For DLO end-tip position control, we use a stiff DLO.

\textbf{Hand-DLO placement}. The hand-DLO placement varies after 10 successes. 
This includes DLO initial angles and hand base positions in Table \ref{table:approach:env_setup}.
We assume the hand's base default position is at the origin of the environment.
The DLO is placed initially $10cm$ above the hand palm.
%, and the DLO's center is aligned with the hand palm.
For each task, the DLO is placed at different positions compared to the hand, see Section \ref{sec:approach:problem}, but the hand still has a random relative shift compared to the default position.
The DLO initial angles are the initial angles of each spherical joint in the DLO.
The default angles are zero, so the DLO is placed as a straight line above the hand.
%We illustrate this in detail in Section \ref{sec:exp:setup}.
The floor height is set to $-0.12m$ in DLO grabbing and DLO end-tip position control and $-0.6m$ in DLO pulling and bending.
Adjusted environment setup parameters are uniformly sampled from the range after 10  successes.
We used the same adaptive scheme as updating DLO parameters to update the randomization ranges of hand-DLO placement.
In addition, we have a special setting for DLO pulling that we initialize the weight to $0.01kg$ and add $0.01kg$ to it after each success until $0.4kg$.

\begin{table}[t!]
\centering
\caption{Std of Gaussian Noise Added to Observation and Action.}
\begin{tabular}{cc}
\hline
Observation \& Action Noise           & $\sigma$             \\ \hline
Hand Joint Position (rad)   & 0.02              \\
Hand Joint Velocity (rad/s) & 0.02              \\ 
DLO Keypoints Position (mm) & 10                \\
Fingertip Position (mm)     & 3                 \\
Action (rad)                & 0.1              \\ \hline
\end{tabular}
\label{table:approach:obs_noise}
\end{table}

\textbf{Noisy observation}. Noisy observation is important for Sim-to-Real implementation in the future. In Table \ref{table:approach:obs_noise}, we show the variance of the observation noise of each type. We follow a similar strategy for randomizing DLO parameters and environment setup parameters to the noise. We add noise after 10 successes and linearly increase the noise from $1\%$ to $100\%$ after another 100 successes.

\subsection{Experiment Results}
\label{sec:exp:results}

\begin{figure}[t!]
  \centering
  \includegraphics[width=0.48\textwidth]{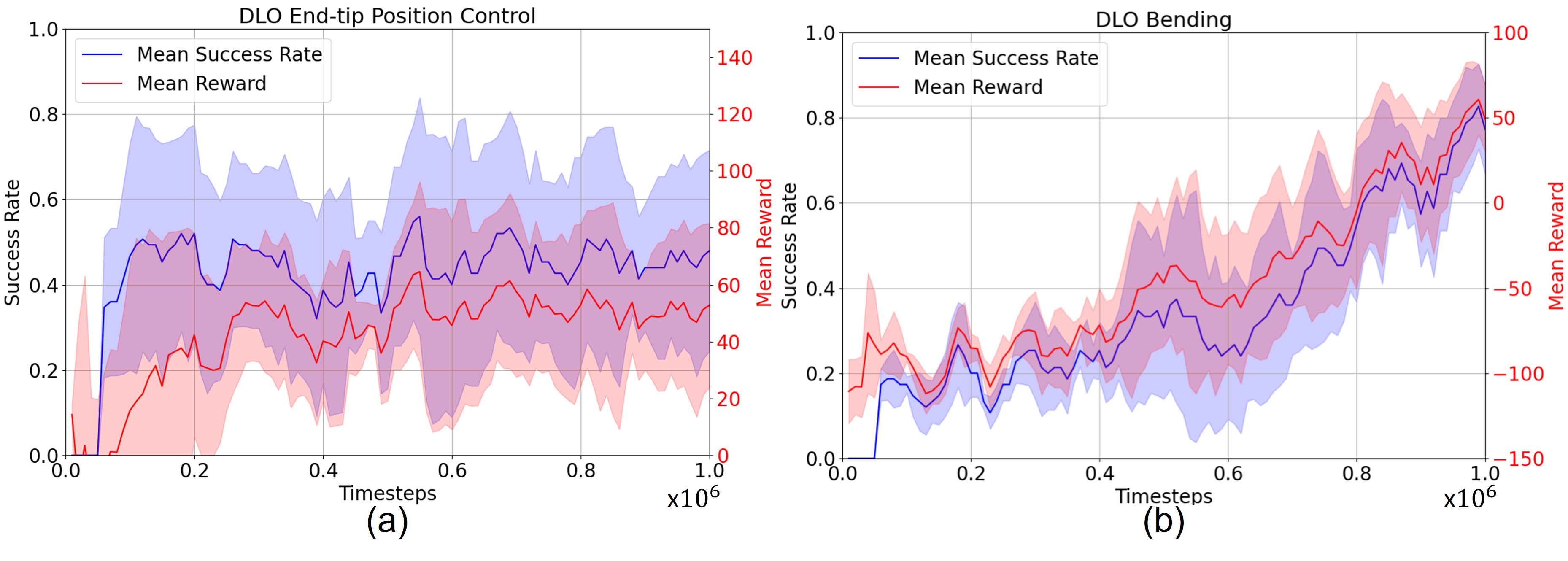}
  
  \caption{Training curves for (a) DLO end-tip position control and (b) DLO bending. We use the same color as Figure \ref{fig:exp:evaluation:top3}.}
  \label{fig:exp:evaluation:lower2}
  \vspace{-3mm}
\end{figure}

Two metrics were used: mean reward per episode during training and success rate of reaching the goal during evaluation. 

The performance of DexDLO was evaluated on three primary tasks 1) DLO grabbing from the right-hand side (DLO grabbing right for short), 2) DLO grabbing from the left-hand side (DLO grabbing left for short), and 3) DLO pulling.
The results are displayed in Figure \ref{fig:exp:evaluation:top3}.
The results show that DexDLO can achieve all three tasks. 
The agent achieved more than $60\%$ success on DLO grabbing right, more than $80\%$ success on DLO grabbing left, and almost $100\%$ success on DLO pulling.
The agents still showed an increasing performance trend when training was terminated.

In addition to performance testing, DexDLO's compatibility and availability are tested on DLO end-tip position control and DLO bending. The results are shown in Figure \ref{fig:exp:evaluation:lower2}. 

DexDLO can also perform DLO bending and DLO end-tip position control, as shown in Figure \ref{fig:exp:evaluation:lower2}.
However, the performance on DLO end-tip position control is not as good as the other four tasks.
DLO end-tip position control curves tend to be flat in the early stages of training, and the end-tip control does not increase in the rest of training.
One possible reason is that our current framework finds it difficult to perform accurate goal-conditioned tasks. Another possible reason is improper goal setting. 
We do not guarantee the goal is reachable, and the controlled point may never reach some randomly sampled goal.

We further tested the speed of the policies. For the DLO grabbing right, it takes $1.9s$ to grab $0.6m$ on average.
For the DLO grabbing left task, it takes $0.8s$ to grab $0.35m$ on average. For the DLO pulling with a $0.4kg$ weight, it takes $13.5s$ to pull $0.3m$ on average.

%Our DexDLO framework can achieve all five goal-conditioned tasks without changing the framework's structure, e.g., observation space and reward functions.
We show that an intuitive framework can solve a complicated control system where the actuators and the manipulated object both have very high DoFs, which is considered extremely challenging and has not been explored before. 

\subsection{Learned Policy Analysis.}
\begin{figure}[t!]
  \centering
  \includegraphics[width=0.44\textwidth]{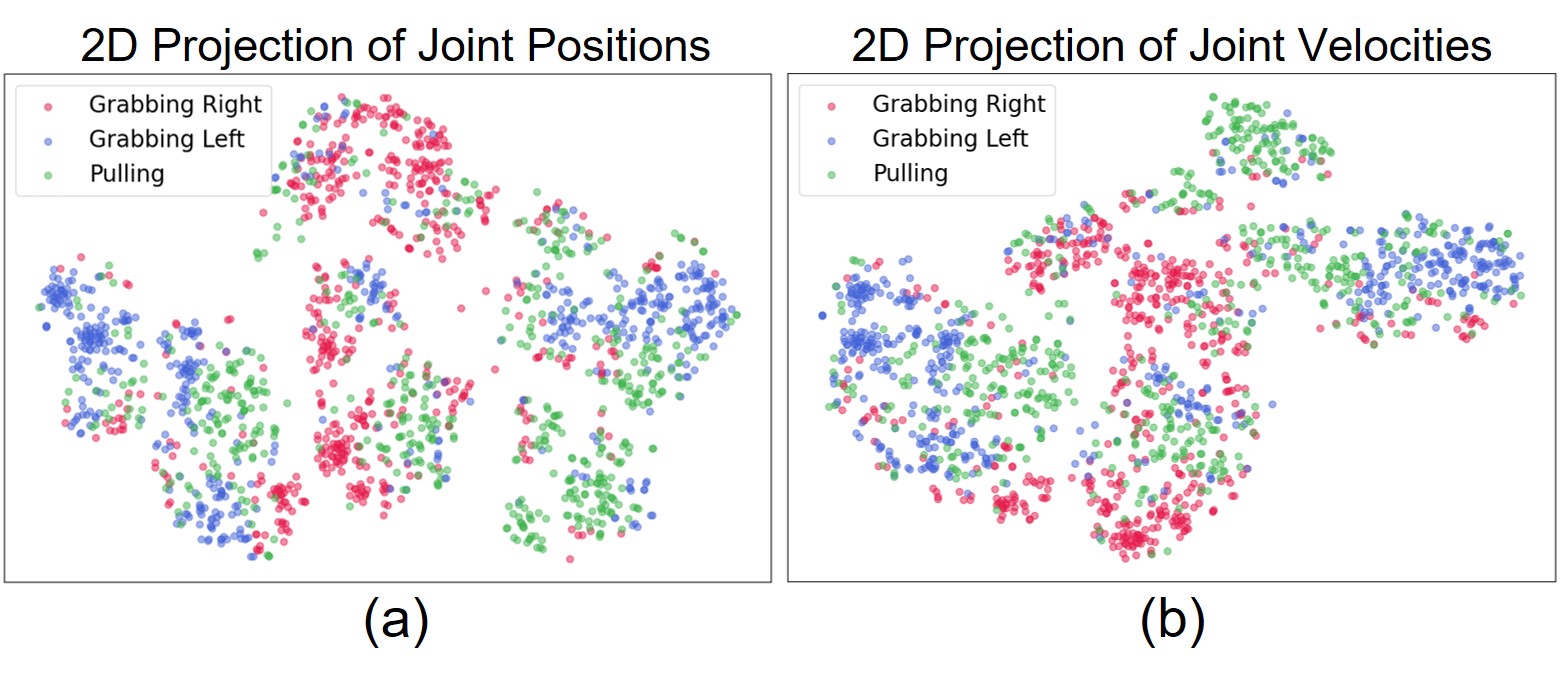}
  \caption{2D projection of hand joint positions (a) and velocities (b) from three tasks by applying t-SNE.}
  \label{fig:exp:skill}
  \vspace{-3mm}
\end{figure}

We further studied if the learned policies of DexDLO on different tasks are similar or not. 
T-SNE \cite{van2008visualizing}, a popular tool to visualize high-dimensional data, was used to analyze three policies for DLO grabbing right, DLO grabbing left, and DLO pulling.
We show the visualization of the 2D projected hand joint positions and hand joint velocities from these three tasks in (a) and (b) of Figure \ref{fig:exp:skill}.
We observe that: 1) Both positions and velocities overlap in small areas for three tasks, indicating a common learned behavior among different tasks. 
2) The two figures show a large separation between DLO grabbing right and DLO grabbing left from the clustered areas. 
The different goals of the two tasks lead to different learned actions (note that the hand is not symmetric).

We speculated that the policies learned for different tasks have a common dexterous manipulation behavior and specialized goal-specific skills.
However, drawing a solid conclusion requires further investigation.

\subsection{Ablation Studies on Reward Function and Observations}
\label{sec:exp:reward}
\begin{figure}[t!]
  \centering
  \includegraphics[width=0.48\textwidth]{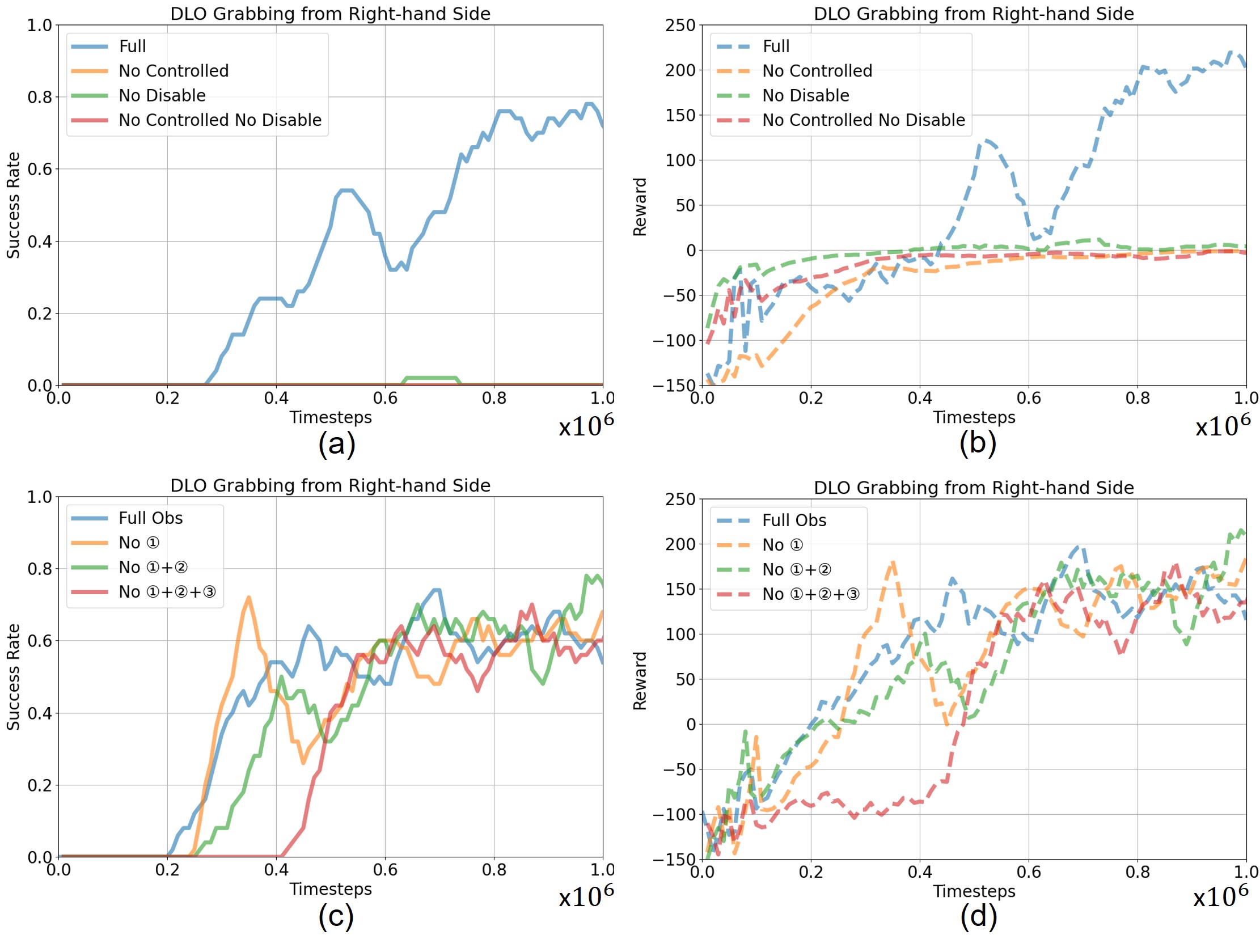}
  \caption{Success rate and reward curves of different reward terms and lack of observations. (a) and (b) show success rate and reward curves with different reward terms. (c) and (d) show success rate and reward curves of removing certain observations. \textit{Full obs} indicates using all observations. \textcircled{\small{1}} indicates hand palm positions and DLO-palm vectors. \textcircled{\small{2}} indicates hand joint velocities. \textcircled{\small{3}} indicates additional 16 keypoints' positions based on existing 9 keypoints. The combination means removing multiple observations together.}
  \label{fig:exp:lack_observation_reward}
  \vspace{-3mm}
\end{figure}

The DLO grabbing right task was used for the ablation studies, including removing certain reward terms and observations.

\textbf{Removing pose-regularized reward terms}. 
We tested the removal of the reward terms $\mathds{1}_{controlled}$ and $R_{disabled}$ which also includes the failure condition.
%on DLO grabbing right.
We show that both reward terms are essential during training the agent in (a) and (b) of Figure \ref{fig:exp:lack_observation_reward}.
By removing one or two pose-regularized reward terms, the agent has a near-zero success rate.
$R_{disabled}$ can help the agent avoid useless explorations. $\mathds{1}_{controlled}$ encourages the agent to learn more efficient manipulation skills by using the thumb and the index finger. 
The agent fails to learn the manipulation task when either reward term is removed.

\textbf{Observation redundancy}. As stated in Section \ref{sec:approach:framework}, we have many observation terms in the observation space, and we want to know if some of them are redundant and can be removed.
We tested using fewer observations: 
1) hand palm positions and DLO-palm vectors (103-d), 
2) hand joint velocities (24-d), 
3) evenly downsampled keypoints' positions (192-d, with only 9 keypoints rather than the full 25 keypoints, which also influences the dimensions of DLO-fingertip positions).
We show three results by gradually removing the observations in (c) and (d) of Figure \ref{fig:exp:lack_observation_reward} with fewer observations, the agent still learned the manipulation policy.
This indicates that these observations are not decisive in the DLO grabbing task, and a potential exists to reduce the observation space in other tasks.

% \subsection{Robustness on Noise}
% \label{sec:exp:robustness}
% \begin{figure}[htbp]
%   \centering
%   \includegraphics[width=0.48\textwidth]{images/experiment_observation_influence.jpg}
%   \caption{Robustness against noise and action Delay. Left: XXX. Right: XXX.}
%   \label{fig:exp:lack_observation_reward}
% \end{figure}

% We further tested if the learned policy is robust against noise in observation and action space and delay input in the action. This is also crucial for future Sim-to-Real implementation.

\section{DISCUSSION AND CONCLUSION}
\label{sec:discussion}

%\subsection{Possibility of finishing a long-horizon task}
%Considering we only select the goal as reaching rather than following, the learned agent has not yet evaluated its ability on multiple goal reaching or goal following. We propose two possible long-horizon tasks, DLO insertion and DLO trajectory following, which may benefit from our goal-conditioned DexDLO.
%\textbf{DLO insertion.} 

%\textbf{DLO trajectory following.}

\subsection{Evidence supporting transfer to the real world}
\label{sec:conclusion:sim2real}
We highlight the importance of the following factors for Sim-to-Real transfer:

1. Observation space design:
Previous works have investigated how to acquire the hand states, like fingertip positions and hand joint angles in the real world \cite{handa2022dextreme}.
As for DLO state estimation, rather than using point clouds or RGBD images, the DLO state in the observation space is DLO keypoints' positions which can be detected and tracked using state-of-the-art algorithms that cope with with occlusions in real-time and in 3D space \cite{lv2023learning, xiang2023trackdlo}. 
No DLO velocities in the observation space are needed. 
Additionally, in Section \ref{sec:exp:reward}, we demonstrated that the agent can effectively learn the policy even when fewer keypoints are available.

%2. Action space design: We use OpenAI's method of the hand's action space. This improves the stability of the manipulation and avoids influence on inaccurate observation of finger joint angles.

2. DLO and dexterous hand simulation: We simulated the DLO in MuJoCo.
The DLO simulated by a series of connected rigid capsule bodies has been used in Sim-to-Real \cite{chi2022iterative, lim2022real2sim2real}
The contact between the DLO and the hand is between rigid objects, which simplifies the physics. 
The Shadow hand model is accurately modeled already \cite{menagerie2022github} with tested realistic parameters. This makes the sim-to-real transfer more likely to proceed feasibly.

3. Randomization in the simulation environment: We randomized the DLO parameters, shown in Table \ref{table:approach:DLO_param}, different environment setups in Table \ref{table:approach:env_setup}, and noise in the observation, shown in Table \ref{table:approach:obs_noise}. 
The ability to tolerate this variation in parameters demonstrates the robustness of our policy and further enhances the sim-to-real capability.

4. Training time: A short training time is important to directly train the agent on real-world robots with minimal wear-and-tear. Nagaband et al. \cite{nagabandi2020deep} implemented their training on the real-world shadow hand with four fours training. Our simulation runs in real-time, around 60 steps per second, matching the speed of the real world. The total training time is $1\times10^6$ steps for each task, which takes less than 5 hours in simulation. 
It is possible to directly train the agent on real-world hardware.

\subsection{Summary}
\label{sec:conclusion:summary}
%[Not finished yet for summary/conclusion]
This paper proposed a general formulation of goal-conditioned DLO dexterous manipulation to summarize a series of common DLO manipulation tasks.
Based on this formulation, we presented a unified framework, DexDLO, that learns goal-conditioned policies to perform a series of dexterous DLO manipulation tasks.
We have evaluated the performance of the learned policies on each task and run ablation studies to validate the importance of proposed pose-regularized reward terms.
%Via the ablation study, we found redundant observations and demonstrated the importance of proposed pose-regularized reward terms.
We discussed the capability and potential of our DexDLO for real-world dexterous DLO manipulations.

%\addtolength{\textheight}{-12cm} 
\bibliographystyle{IEEEtran}
\bibliography{scibib}

\end{document}